\documentclass[preprint,12pt]{elsarticle}

\usepackage{amssymb}
\usepackage{amsmath}

\journal{International Journal of Imaging Systems and Technology}

\usepackage{lineno}
\usepackage{subcaption}
\usepackage{graphicx}
\usepackage{mwe}
\usepackage{adjustbox}
\usepackage{booktabs}

\begin{document}

\begin{frontmatter}

\title{DE-KAN: A Kolmogorov Arnold Network with Dual Encoder for accurate 2D Teeth Segmentation}

\author[label1]{Md Mizanur Rahman Mustakim}
\author[label1,label2]{Jianwu Li\corref{cor1}}
 \author[label3]{Sumya Bhuiyan}
\author[label4]{Mohammad Mehedi Hasan}
\author[label5]{Bing Han}

\affiliation[label1]{organization={School of Computer Science and Technology, Beijing Institute of Technology},
            city={Beijing 100081},
            country={PR China}}
\affiliation[label2]{organization={National Engineering Research Center of Oral Biomaterials and Digital Medical Devices},
            city={Beijing 100081},
            country={PR China}}
\affiliation[label3]{organization={Department of Statistics and Data Science, Jahangirnagar University},
            city={Savar 1342},
            country={Bangladesh}}
\affiliation[label4]{organization={Faculty of Information Technology, Beijing University of Technology},
            city={Beijing 100124},
            country={PR China}}
\affiliation[label5]{organization={Department of Orthodontics, Peking University School and Hospital of Stomatology, National Center for Stomatology, National Clinical Research Center for Oral Diseases, National Engineering Research Center of Oral Biomaterials and Digital Medical Devices, Beijing Key Laboratory of Digital Stomatology, NHC Key Laboratory of Digital Stomatology, NMPA Key Laboratory for Dental Materials, No.22, Zhongguancun South Avenue, Haidian District, Beijing, 100081, PR China}}

\cortext[cor1]{{Corresponding Author: Jianwu Li}, {email: ljw@bit.edu.cn}}

\begin{abstract}
Accurate segmentation of individual teeth from panoramic radiographs remains a challenging task due to anatomical variations, irregular tooth shapes, and overlapping structures. These complexities often limit the performance of conventional deep learning models. To address this, we propose DE-KAN, a novel Dual Encoder Kolmogorov Arnold Network, which enhances feature representation and segmentation precision. The framework employs a ResNet-18 encoder for augmented inputs and a customized CNN encoder for original inputs, enabling the complementary extraction of global and local spatial features. These features are fused through KAN-based bottleneck layers, incorporating nonlinear learnable activation functions derived from the Kolmogorov Arnold representation theorem to improve learning capacity and interpretability. Extensive experiments on two benchmark dental X-ray datasets demonstrate that DE-KAN outperforms state-of-the-art segmentation models, achieving mIoU of 94.5\%, Dice coefficient of 97.1\%, accuracy of 98.91\%, and recall of 97.36\%, representing up to +4.7\% improvement in Dice compared to existing methods. 
The source code is available at https://github.com/MizanMustakim /DEKAN.

\end{abstract}



\end{frontmatter}



\section{Introduction}

In human physiology, teeth significantly influence food decomposition, digestion, and nutrient assimilation. Healthy teeth are essential for optimal dietary function, overall health and well-being. They also affect eating, digestion, voice clarity, face shape, and general wellness. Due to these multifarious functions, dental imaging has become essential to clinical practice, facilitating diagnostics, treatment planning, and post-procedure evaluations. As a result, imaging in dentistry can offer clinicians the critical data needed to make informed decisions and take further action in patient care, including detecting cavities, diagnosing structural abnormalities, and planning orthodontic treatments \cite{1}. Traditionally, dentists and radiologists manually analyze these images, depending on their experience, to detect abnormalities, including dental caries, structural skeletal abnormalities, and misalignment of teeth \cite{2,3,4,5}.

\begin{figure}[t]
   \begin{center}
   
   \begin{subfigure}{0.5\textwidth}
   \begin{center}
      \includegraphics[width=0.9\linewidth]{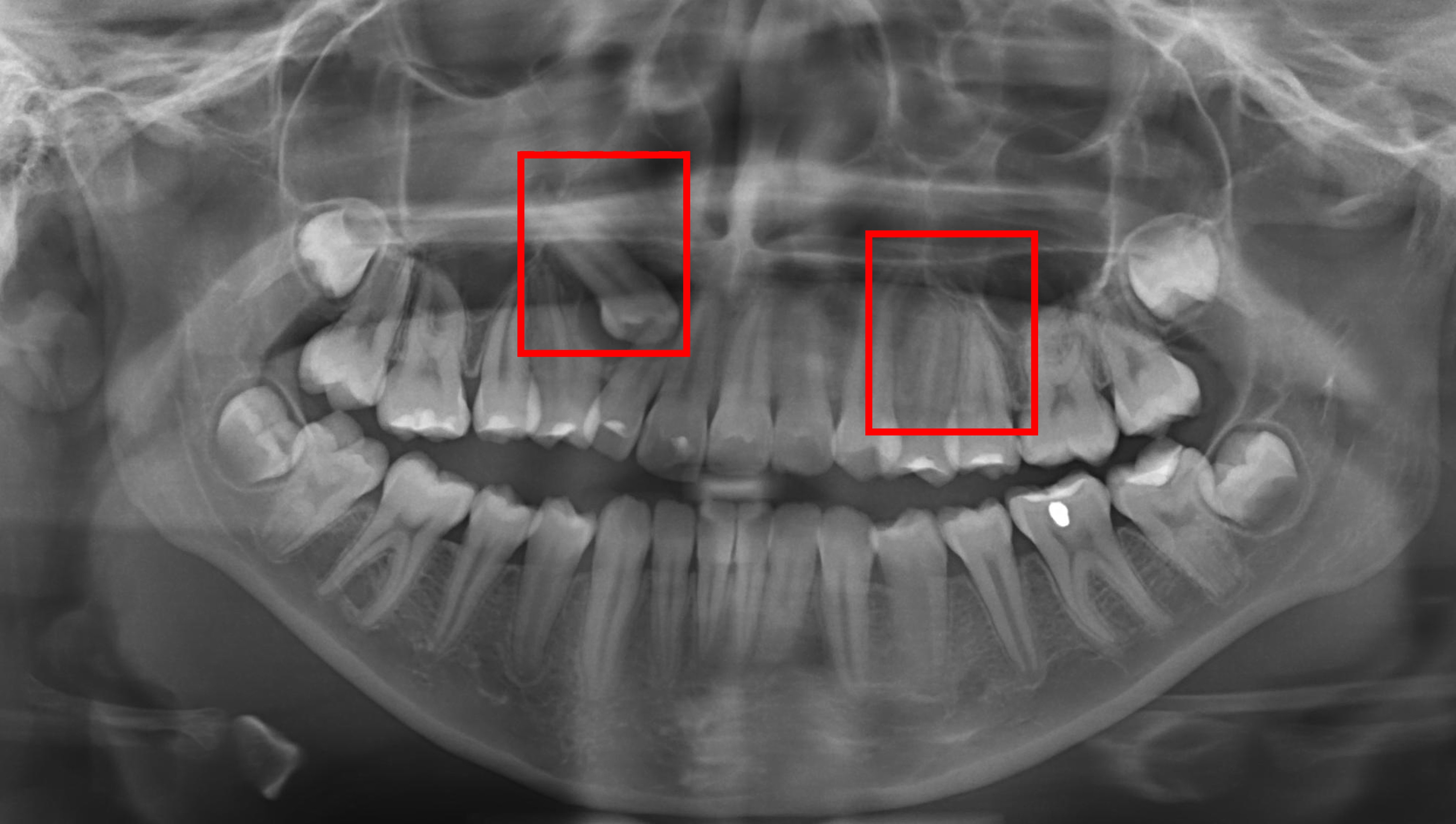}
      \caption{}
      \label{fig:1(a)}
    \end{center}
   \end{subfigure}%
   \begin{subfigure}{0.5\textwidth}
   \begin{center}
      \includegraphics[width=0.9\linewidth]{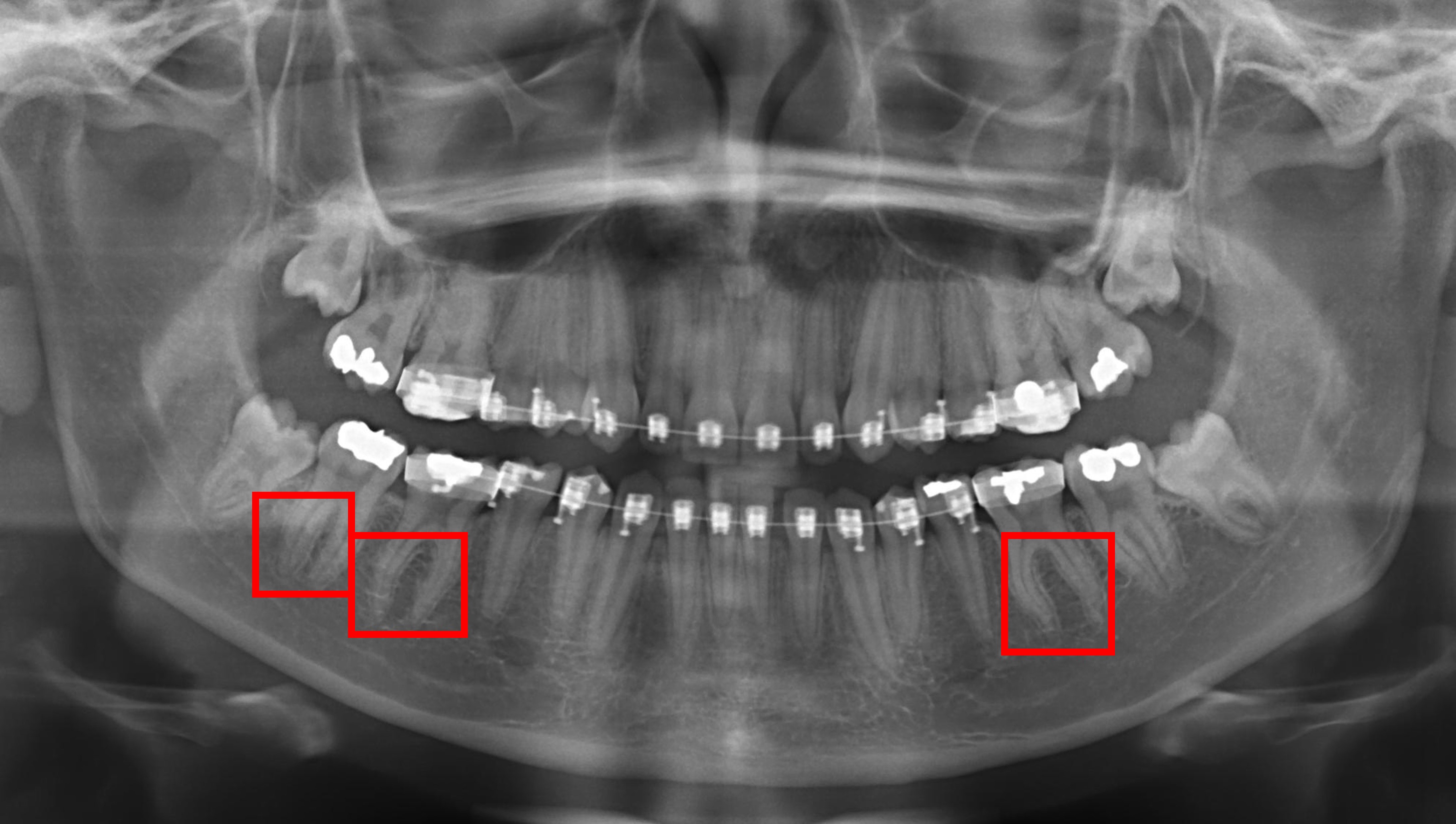}
      \caption{}
      \label{fig:1(b)}
   \end{center}
   \end{subfigure}

   \caption{(a) shows the overlapped teeth, and (b) denotes the sharpness of teeth in different shapes. 
   \label{fig:1} 
   }  
   \end{center}
\end{figure}

However, manual interpretation has significant areas for improvement: it is time-consuming, susceptible to human error, and highly reliant on the clinician's competence. Separating and identifying various components of dental structures, including teeth, roots, and adjacent tissues, within medical images (e.g., X-rays, MRI, and CT scans), is one of the most critical areas of focus in dental imaging. According to Zhang et al.\cite{6}, segmentation divides an image into meaningful regions corresponding to anatomical features. It is increasingly necessary to implement automatic and precise segmentation of dental images. The underpinning of this urgency is an abundance of variables. Manual interpretation may be subjective initially; in equivocal situations, two specialists may not consistently agree on the exact diagnosis or treatment approach. As illustrated in Figure~\ref{fig:1(b)}, the complex character of dental structures and the proximity of tangible elements, such as teeth and jawbones, present substantial obstacles. Finally, human eyes are impaired in their ability to differentiate between teeth in dental X-rays due to the frequent overlap of teeth, as illustrated in Figure \ref{fig:1(a)}. Identifying the fine details of dental roots and cavities may prove challenging in low-resolution images. The existence of these obstacles underscores the urgent need to automate these procedures with more resilient, precise, and scalable methodologies.

To address these challenges, previous work has explored automated dental image segmentation using advanced computational techniques, including traditional machine learning (ML) and deep learning (DL). Traditional image processing methods, such as edge detection \cite{7}, thresholding \cite{8}, and region-based methods \cite{9}, make it challenging for early methods to deal with complex dental structures. Recent advances in DL, particularly Convolutional Neural Networks (CNNs), such as Mask-R-CNN, PANet, HTC, and ResNet, have revolutionized the field by leveraging hierarchical feature extraction and end-to-end learning capabilities in tooth segmentation \cite{10}. Despite these advancements, handling the pixels of sharp-edged objects, such as damaged teeth, remains challenging, and sometimes, during scanning, the presence of objects is not clearly visible in dental imaging. Traditional methods only enhance the single encoder for feature extraction so that it is hard for them to deal with these issues when data limitations are challenged.

To challenge these issues, we propose a dual encoder network, inspired by Zou et al.\cite{11}, that integrates image augmentation \cite{12} and non-augmentation (original image) strategies, addressing the limitations of traditional encoder architectures for precise tooth segmentation. In contrast to traditional single-stream networks, such as Mask R-CNN \cite{mask_rcnn} or U-Net \cite{unet}, our approach uses two separate encoders. One encoder is based on a ResNet-18 \cite{resnet} backbone that is designed for hierarchical abstraction from significantly augmented input images, while the other uses a customized CNN to capture detailed extraction of local features from the original input images. This dual input method allows the network to effectively capture a wider range of dental features, making segmentation more accurate and reliable across a wider range of image quality and structural complexity and preserving intricate information about input images.

Current deep learning models in medical image segmentation encounter difficulties arising from inadequate kernel design and insufficient explainability \cite{16}. Traditional methodologies, such as convolutional operation \cite{17}, Transformers \cite{transformer}, and Multi-Layer Perceptrons (MLPs) \cite{MLP}, address spatial dependencies linearly across channels, restricting their ability to represent complex non-linear patterns frequently encountered in medical imaging \cite{20}. These patterns often exhibit diverse clinical significance, reflecting various anatomical or pathological characteristics \cite{21}. The empirical and heuristic characteristics of these methods compromise interpretability and explainability, thus diminishing trust in clinical decision making. Spatial and contextual information is easily lost by these methods, degrading their effect of segmentation in complicated situations.

Thus, we present a module in the bottleneck of our architecture using Kolmogorov Arnold Networks (KANs) introduced by Liu et al.\cite{KAN}, which are more successful in complex image scenarios. KANs use the Kolmogorov-Arnold theorem for decomposing multivariate functions into interpretable univariate components, providing a "white-box" approach that improves transparency. KANs can improve feature optimization and generalization by processing features derived from the dual encoder in our proposed method. This provides robust oversight for overlapping structures, low-contrast areas, and hazy borders between teeth and jawbones. This attribute makes KANs more advantageous, particularly in scenarios requiring exceptional reliability and accuracy in real-world tasks. Our main contributions are summarized as follows.

\begin{enumerate}
    \item We propose a dual input architecture that uses image augmentation techniques to avoid losing information about input images during segmentation. It employs two distinct feature extraction encoders: a pre-trained ResNet-18 for strongly augmented images and a customized CNN for original images, ensuring diverse and comprehensive feature learning.

    \item We introduce a combination of two KAN blocks as the bottleneck, incorporating KANs for hierarchical and interpretable feature refinement. We design a novel hybrid approach in the KAN blocks that seamlessly integrates KANs into traditional CNN layers, enhancing their interpretability and feature optimization capabilities.

    \item Our proposed model outperforms state of the art segmentation approaches on two dental panoramic X-ray benchmark datasets.
\end{enumerate}

\label{sec1}

\section{Related work}
\label{sec2}
\subsection{Dental Image Segmentation}
\label{sec2.1}
Deep learning methodologies, particularly CNNs \cite{23} and Transformers \cite{transformer}, have significantly enhanced the precision and reliability of 2D dental image segmentation. These methods are adept at generating teeth-specific feature maps and improving segmentation accuracy \cite{24}. The well-known architectures U-Net \cite{unet} and V-Net \cite{25} address primarily medical image segmentation problems. Koch et al.\cite{26} employed U-Net with completely convolutional neural networks to perform semantic segmentation on dental panoramic radiographs. To improve U-Net, Hou et al.\cite{27} added a squeeze-excitation (SE) module \cite{28} to the encoder and swapped out the bottleneck for a multi-scale aggregation attention block (MAB). This enhances its ability to manage irregular teeth. To address challenges such as identifying damaged or overlapping teeth, Almalki et al.\cite{29} introduced a residual U-Net based on a denoized encoder, focusing on specific regions within the images to improve segmentation accuracy. Similarly, Li et al.\cite{30} proposed a Transformer-based hybrid U-Net architecture, replacing conventional encoders and decoders with a group Transformer. This hybrid structure combines CNNs and Transformer-based designs, making it much easier to find edges and more accurate when cutting tooth roots out of jawbones. 
Supervised learning approaches, particularly U-Net variations, have consistently delivered state-of-the-art performance in dental 2D segmentation. However, a persistent challenge remains the need for labeled dental 2D datasets. To handle this, Sun et al.\cite{31} proposed the semi-supervised tooth segmentation Transformer U-Net (STS-TransUNet), incorporating spatial and channel attention mechanisms in the decoder to effectively leverage labeled and unlabeled data while capturing detailed feature dependencies. In addition to these Transformers- and U-Net-based methods, CNN-based architectures have also demonstrated success in dental panoramic X-ray segmentation. For example, Muresan et al.\cite{32} introduced the Efficient Residual Factorized Convolutional Network (ERFNet), an encoder-decoder model with 23 convolutional layers. ERFNet mitigates degradation problems by employing residual functions, ensuring accurate and stable segmentation results. These advancements underscore the various strategies used to overcome challenges in dental image segmentation.

Another CNN-based model, "You Only Look Once" (YOLO), has become a leading model for object detection, segmentation, and classification tasks \cite{yolo}. Deepho et al.\cite{34} showed that YOLOv8 outperforms previous YOLO iterations in tooth segmentation tasks, improving both accuracy and efficiency. However, the complex nature of dental panoramic X-ray images necessitates the use of hybrid approaches to capture deep feature representations. Zhao et al.\cite{35} developed a Two-Stage Attention Mechanism Model (TSASNet), which improves feature extraction and segmentation at the pixel level by utilizing global and local contexts in panoramic X-rays. Despite its progress, TSASNet faces challenges in accurately segmenting foreground pixels into separate dental areas. To address this problem, Chen et al. \cite{36} introduced the Multiscale Location Perception Network (MSLPNet), using ResNet-50\cite{resnet} to extract local and global information across several scales. They improved accuracy by implementing a hybrid loss function that integrates MS-SSIM loss, Dice loss, and binary cross-entropy (BCE) loss. Despite MSLPNet's superior optimization, its intricate design and constraints in segmenting overlapping teeth render it less effective on varied datasets. To address the complexities of overlapping and varied tooth segmentation, Chandrashekar et al.\cite{37} devised a collaborative learning methodology that enhances Mask R-CNN \cite{mask_rcnn} and Faster R-CNN \cite{38} models to boost the localization and segmentation of individual teeth. Yosinski et al.\cite{39} have acknowledged transfer learning as an effective method for ensuring consistency in feature extraction across images of comparable dimensions.

Although the aforementioned methods have demonstrated promising results, there is potential feature loss in the encoders of existing models. The application of data augmentation techniques in a single stream encoder cannot mitigate the risk of losing features from dental images during feature extraction, as each corner of the feature set is essential due to the limitations of medical data. In order to resolve this matter, we implement a Dual-Encoder feature extractor in our network. This involves the integration of customized tiny CNNs to extract local spatial features from original images and a backbone of ResNet-18\cite{resnet} to extract highly abstracted global features under various conditions from enormously augmented images.

\subsection{Kolmogorov Arnold Network}
\label{sec2.2}
\begin{figure}[h]
  \centering
  \includegraphics[width=0.7\linewidth]{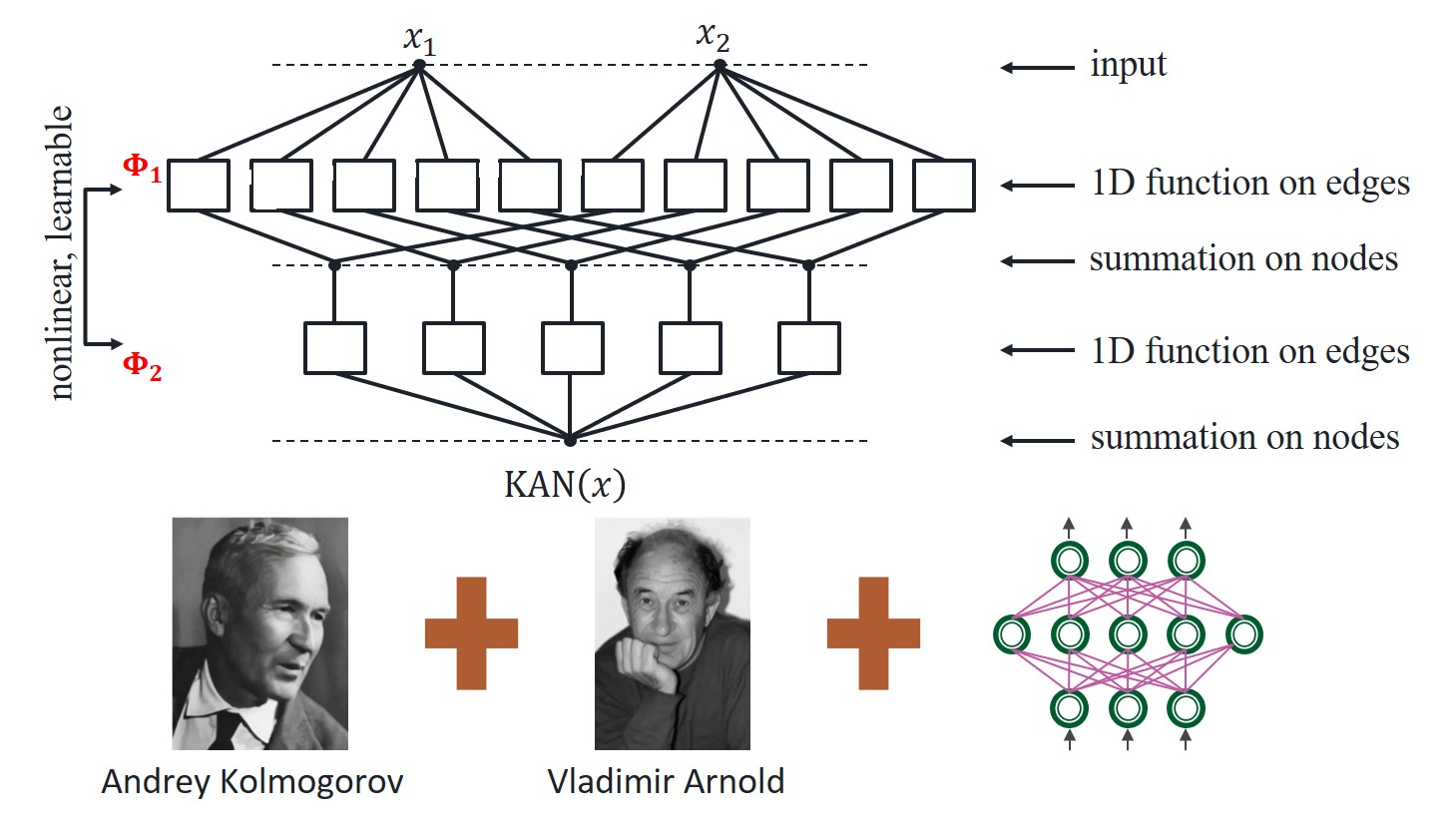}
  \caption{A visual representation of Kolmogorov Arnold Network.}
  \label{fig:1.2}  
\end{figure}
The notion of Kolmogorov Arnold Networks (KANs), proposed by Liu et al.\cite{KAN}, is derived from the Kolmogorov Arnold Representation (KAR) theory, a comprehensive mathematical framework that encompasses universal approximation, dynamical systems theory, and non-linear transformations \cite{40,41}. As visual in Figure~\ref{fig:1.2} this theory underpins KANs, allowing them to acquire intricate data representations via nonlinear transformations in elevated-dimensional feature spaces. This representation theory can be expressed as:
\begin{equation}
  f(x) = f(x_1, \dots, x_n) = \sum_{q=1}^{2n+1} \Phi_q \left( \sum_{p=1}^{n} \phi_{q,p}(x_p) \right), 
  \label{eq1}
\end{equation}
where $\Phi_q$ and $\phi_{q,p}$ are univariate functions, $\phi_{q,p}:\left[0,1\right] \to \mathbb{R}$ and $\Phi_{q}:\mathbb{R}\to\mathbb{R}$. Before KANs, Igelnik et al.\cite{42} presented Kolmogorov's Spline Network (KSN), which uses KAR theory to improve data adaption. KSN is good at multivariate function approximation, especially for datasets with many dimensions and few elements, because it uses cubic spline methods for activation functions and internal representations. Notwithstanding its advantages, KSN encounters several constraints. The network's intricacy leads to excessive spline segments, rendering the calculations for back-propagation and gradient descent impractical. This necessitates specialized methods for spline fitting and piecewise polynomial optimization, which presents practical challenges.

Liu et al. introduced KANs to address limitations in existing models, enabling nonlinear transformations at every layer to improve computational efficiency and generalization \cite{KAN}. KANs have been successfully applied to various tasks, including recommendation systems \cite{43}, real-time satellite traffic forecasting \cite{44}, and complex mechanical problems \cite{45}, consistently outperforming traditional models like MLPs in accuracy and efficiency. KANs have shown significant potential in medical image segmentation, surpassing traditional MLP-based methods. Li et al.\cite{46} achieved notable advancement by presenting an advanced network known as U-KAN. This hybrid design combines the traditional U-Net model with KANs, incorporating tokenized KAN layers in both the encoder and decoder, which are connected by a bottleneck. Their research shows that adding KANs to U-Net significantly improves segmentation accuracy, making it work better than regular U-Net and its variations, such as TransUNet \cite{47}. This development underscores the potential of KANs to transform medical image segmentation by delivering more accurate and efficient models adept at managing complex medical data, including dental X-rays and CT scans.

Motivated by these existing methodologies, we present a KAN block that combines KAN linear layers and CNNs at the bottleneck. KAN blocks help mitigate the risk of feature loss after feature extraction in our network, as each extracted feature is essential for precise tooth segmentation, particularly when addressing pointed edges of teeth and overlapped teeth. KANs are adept at segmenting medical images due to their ability to detect intricate spatial hierarchies and subtleties. This makes them particularly advantageous for tasks such as dental image segmentation, which requires precision for diagnosis and treatment planning. The subsequent chapter provides a thorough elucidation of our methodology and additional analysis.

\section{Methods}
\label{sec3}

\subsection{Preprocessing}
\label{sec3.1}

During training, the proposed DE-KAN framework processes two types of input, augmented and non-augmented images, through separate encoders. This dual input design, combined with data augmentation techniques \cite{48}, enhances the robustness of the model by introducing variability into the training process. One encoder focuses on augmented images to address challenging scenarios, such as low-quality or ambiguous features, while the other extracts features from original inputs to preserve unaltered information. This results in different weight configurations optimized for different data characteristics.

In real-world scenarios, medical images may exhibit reduced visibility or decreased sharpness, which can hinder the distinction between teeth and jawbone or complicate the identification of overlapping teeth. To introduce these scenarios into the model, we apply \textit{RandomBrightnessContrast} to simulate varying illumination, \textit{GaussianBlur} to mimic low-quality imaging, and \textit{HueSaturationValue} to replicate diverse color conditions, all with probability 100\% and processed by the ResNet-18 encoder in our network. We implement these augmentations using the Albumentations \footnote{https://github.com/albumentations-team/albumentations} Python library. The network's 2-to-1 input-output configuration limits the use of certain augmentation methods, such as random cropping \cite{49}, which may cause mismatches between the augmented input and the corresponding target mask. During testing, both encoders process identical and unaltered inputs to ensure consistent and robust feature extraction. The encoder trained on augmented data is particularly adept at extracting meaningful features in visually challenging scenarios without altering the input. The extracted features from both encoders are fused and passed to the bottleneck for further processing, ensuring a comprehensive feature representation for reliable performance in deployment.

\subsection{Model architecture}
\label{sec3.2}
\begin{figure}[t]
  \centering
  \includegraphics[width=\linewidth]{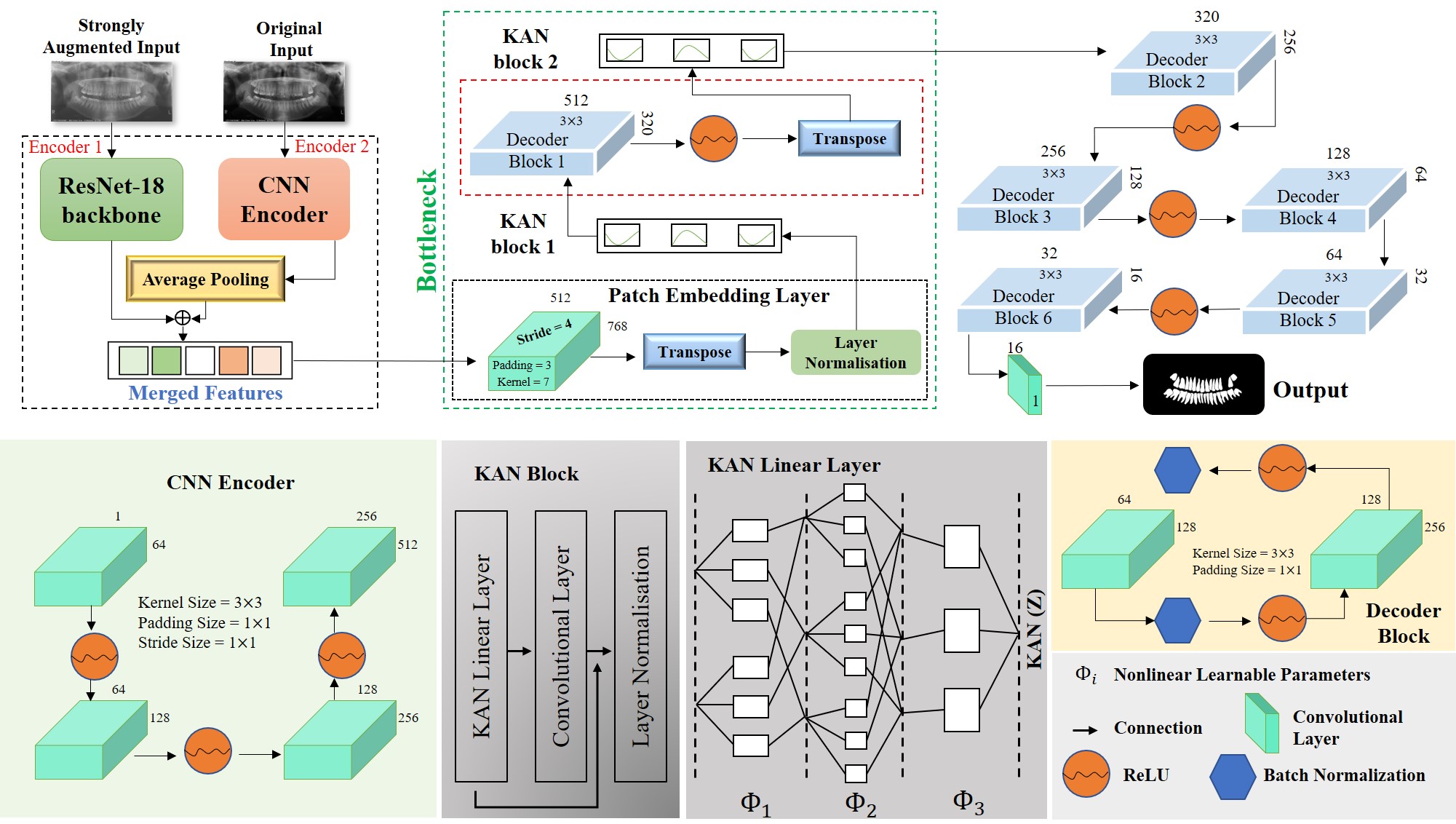}
  \caption{Overview of proposed segmentation network DE-KAN, integrating dual-encoder and KAN blocks. The upper portion illustrates the overall workflow, where the ResNet-18 backbone processes strongly augmented inputs and the CNN extractor processes original inputs, followed by merged feature representations. The hybrid bottleneck incorporates two KAN blocks, including Patch Embedding Layer, KAN Linear Layer, and convolutional components for optimized feature extraction. The decoder reconstructs the segmentation map using hierarchical up-sampling blocks. The lower portion provides detailed views of the CNN encoder, the KAN block architecture, the KAN linear layer with non-linear mappings, and the decoder design, emphasizing the role of non-linear learnable parameters and convolutional layers in pixel-level segmentation.}
  \label{fig:2}  
\end{figure}

In dental segmentation, the exact detection of sharp edges of teeth, roots, and jawbones, together with the extraction of comprehensive contextual and spatial information, is crucial for correctly identifying individual teeth. As seen in Figure \ref{fig:2}, our proposed model proficiently emphasizes the pronounced edges of the teeth while accurately capturing complex spatial details by efficiently extracting and using local-global data from dental x-ray images. Our model can be divided into three sections: the feature extractor module, the bottleneck module, and the decoder module.

\textit{1. Feature extractor modules:} 
This module incorporates a ResNet-18 backbone and a customized 4-layer CNN for down-sampling. In the network, the ResNet-18 encoder processes augmented images, while the customized CNN encoder handles non-augmented (original) images. During training, the ResNet-18 encoder is specifically trained on augmented images to capture various features, such as poorly visible teeth and jawbones. However, no augmentation is required during testing, ensuring a straightforward application. The customized CNN encoder processes nonaugmented images to extract local features. It employs four convolutional layers with progressively increasing feature maps (from 64 to 512), allowing it to learn hierarchical representations. The size of the kernel $3\times 3$, the padding of 1, and the stride of 1 ensure that local spatial patterns, such as edges and textures, are captured in the previous layers. As the network progresses, larger receptive fields in the deeper layers enable the aggregation of spatial information across larger image regions. This structured hierarchy maintains spatial details without grouping too many features. This process ensures that local features, such as fine textures and spatial relationships, which are crucial to segmenting teeth, remain consistent at all stages. Formally, for a given input image $x$, the feature map in the $n^\text{th}$ layer is defined as:
\begin{equation}
  f_n(x) = \sigma(W_n*f_{n-1}(x)+b_n),
  \label{eq2}
\end{equation}
where $W_n$ represents the weights of the $n^\text{th}$ convolutional layer, $*$ denotes the convolution operation, $b_n$ is the bias and $\sigma$ is the activation function.

\begin{figure}[t]
  \centering
  \includegraphics[width=0.6\linewidth]{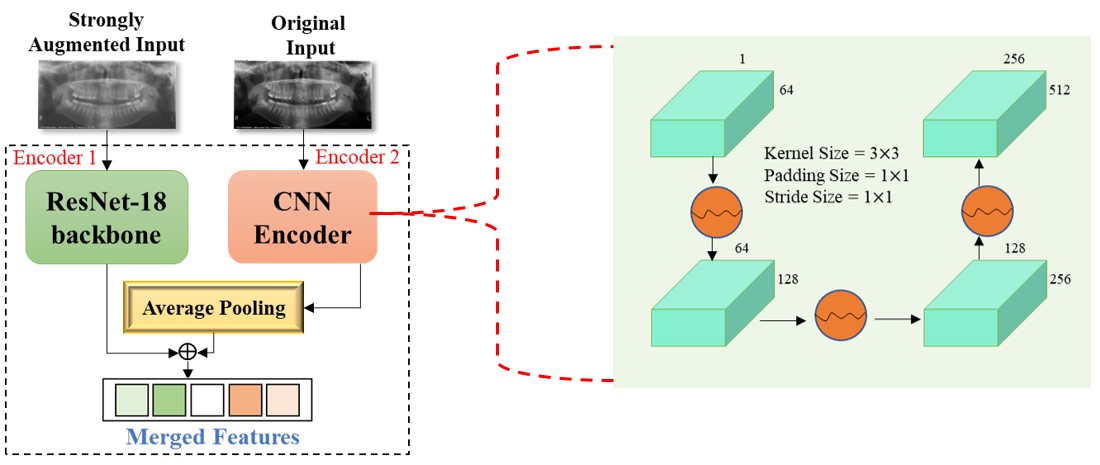}
  \caption{A visualization of the feature extractor module.}
  \label{fig:2.1}  
\end{figure}

Conversely, the ResNet-18 encoder, a pre-trained network, processes augmented images to extract hierarchical abstract features. With its residual connections, depth, and stride-2 convolution with pooling layers, the ResNet-18 encoder looks at the whole input image and finds a broad spatial context and semantic patterns. This encoder is particularly advantageous for dental segmentation tasks, as it enables the model to understand nonuniform tooth shapes, sharp edges, and complex spatial structures by analyzing large receptive fields. As shown in Figure~\ref{fig:2.1}, the features extracted by these two encoders, ResNet-18 for augmented images and the customized CNN encoder for nonaugmented images, are merged to enhance pixel-wise information before passing through the bottleneck. This merged output combines spatial details and semantic abstractions, enriching the model's understanding of intricate tooth patterns. We can express the merging process as the element-wise summation of the feature representations extracted by the ResNet-18 and customized CNN encoders. For an augmented input image $x_a$ and a nonaugmented input image $x_n$, the merged feature map $F_\text{merged}$ is formally defined as:
\begin{equation}
 F_\text{merged} = F_\text{ResNet}(x_a)+AdaptiveAvgPooling(F_\text{CNN}(x_n)),
 \label{eq3}
\end{equation}
where $F_\text{ResNet}(x_a)$ presents the abstract hierarchical features of the ResNet-18 encoder and $F_\text{CNN}(x_n)$ corresponds to the spatially detailed local features from the customized CNN encoder. \textit{AdaptiveAvgPooling} ensures the alignment of the dimensions of both feature maps for effective merging. The fused feature map integrates detailed spatial features from the customized CNN encoder with high-level semantic features from the ResNet-18 encoder, facilitating accurate segmentation of complex and diverse dental structures. In contrast to single-stream encoders, our dual-encoder method enhances feature representation by utilizing identical extractions, thereby improving segmentation performance without any loss of features.

\textit{2. Bottleneck module:} 
The bottleneck module is essential in our architecture, located after the feature extractor. Incorporating KANs allows us to surpass conventional methods in the bottleneck, such as CNNs \cite{23}, MLPs \cite{MLP}, and Transformers \cite{transformer}, by maintaining the structural relationships within the data, which are crucial for accurate decoding. Moreover, KANs improve the interpretability of models, allowing a deeper understanding of their behavior and aiding in the optimization process.
\begin{figure}[h] 
 \centering 
 \includegraphics[width=\textwidth]{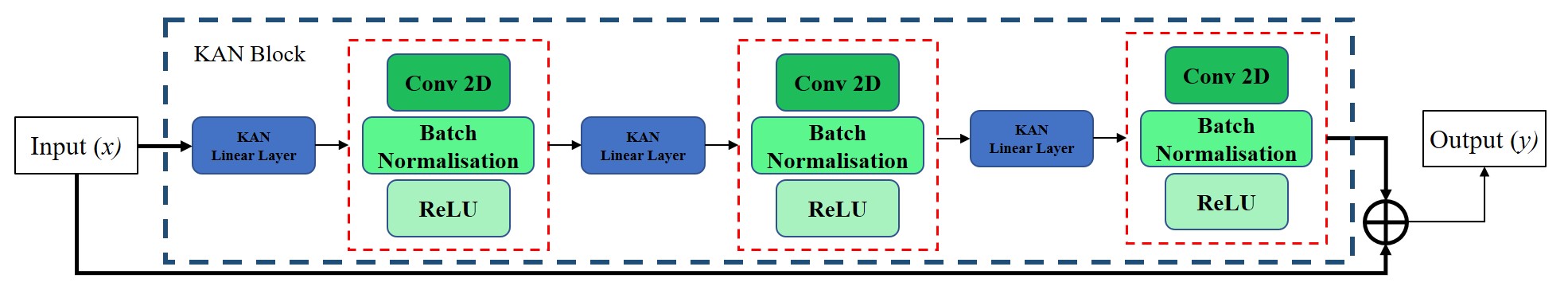}
 \caption{Block diagram of KAN Block, showcasing sequential operations within KAN Linear layers interleaved with convolutional blocks (Conv2D, Batch Normalization, and ReLU activation). The input $x$ undergoes transformation through these stages, and a residual connection ensures the combination of input with the final output $y$, enhancing learning efficiency and stability.} 
 \label{fig:3}
\end{figure} 

Our bottleneck consists of two KAN blocks: KAN block-1 and KAN block-2. As shown in Figure~\ref{fig:3}, each block is made up of three convolutional KAN layers \cite{50}. By embedding learnable nonlinear activation functions of KAN into convolutional operations, this design improves feature extraction while preserving complex spatial and contextual information. In KAN block-1, it processes $F_\text{merged}$ as input through a patch embedding layer, converting 2D spatial features into a flattened sequence of patches. This transformation effectively captures local and global relationships. Formally, for an input tensor $x \in \mathbb{R}^{B \times C \times H \times W}$, where $B$ is the batch size, $C$ is the number of channels, and $H$ and $W$ are the spatial dimensions, the patch embedding layer is applied.
\begin{equation} 
 x_\text{proj} = \Gamma(x, W_\text{proj}, b_\text{proj}), 
 \label{eq4} 
\end{equation} 
where $W_\text{proj}$ and $b_\text{proj}$ represent the weights and biases of the convolutional projection layer $\Gamma$, respectively. This output is reshaped into a sequence: 
\begin{equation} 
 x_\text{patches} = \text{Flatten}(x_\text{proj}) \in \mathbb{R}^{B \times N \times D}, 
 \label{eq5} 
\end{equation} 
where $N = \frac{H}{P} \times \frac{W}{P}$ is the number of patches, $P$ is the patch size, and $D$ is the embedding dimension. Layer normalization ($\mathcal{L}$) is applied to stabilize the training:
\begin{equation} 
 x_\mathcal{L} = \mathcal{L} (x_\text{patches}) 
 \label{eq6} 
\end{equation}
The normalized output $x_\mathcal{L}$ is fed into KAN block-1. Three convolutional KAN layers, modeled as follows, compose the KAN blocks: 
\begin{equation} 
  \text{K}_{\text{block}} = \prod_{i=1}^{3} \Gamma_i \circ \kappa_i(x_\mathcal{L}), 
  \label{eq7} 
\end{equation} 
where $\kappa$ stands for KAN linear layers and $\Gamma$ for convolutional 2D layers with \textit{batch normalization} and \textit{ReLU} activation function. The KAN linear layers are as follows: 
\begin{equation} 
  \kappa(x) = \left(\Phi_{3} \circ \Phi_{2} \circ \Phi_{1}\right)(x), 
  \label{eq8} 
\end{equation}  
where each KAN linear layer $\kappa$ uses a $n_{in}(=2)$-dimensional input and $n_{out}(=2)$-dimensional output, $\Phi$ is defined as: 
\begin{equation} 
  \Phi = \left\{\phi_{q,p}\ \middle|\ p = 1, 2; \ q = 1, 2 \right\}, 
  \label{eq9} 
\end{equation}
where $\phi$ represents learnable nonlinear activation functions. Consequently, we can formally express KAN block-1 as follows: 
\begin{equation} 
  K1_{out} = x_\mathcal{L} \oplus \text{K}_\text{block}, 
  \label{eq10} 
\end{equation} 
where $\oplus$ denotes a residual connection. This ensures stable training and improved convergence. In KAN block-2, the normalized output of KAN block-1 is passed to the first upsampling layer of the decoder, which is then fed into KAN block-2. This block follows the same processing pipeline as KAN block-1, ensuring consistent feature refinement and propagation. The overall bottleneck can be formally expressed as follows: 
\begin{equation} 
  F_{bottleneck} = \mathcal{L} (K1_{out}\circ \Gamma_{\mathcal{U}} ) \oplus \left(K1_{out} \circ \Gamma_{\mathcal{U}1} \circ \text{K}_\text{block}\right), 
  \label{eq11} 
\end{equation}
where $F_{bottleneck}$ is the features of our bottleneck, $\mathcal{L}$ is layer normalization, and $\Gamma_{\mathcal{U}1}$ is the first upsampled convolutional layer of the decoder module.

\textit{3. Decoder module:}
For the segmentation task, we design the decoder module to progressively reconstruct the spatial resolution of the feature maps \( F_\text{bottleneck} \) generated by the bottleneck. This process involves sequential upsampling and feature refinement. Formally, for a feature map \( F \in \mathbb{R}^{C \times H \times W} \), each decoder stage doubles the spatial resolution using bilinear interpolation, producing \( F_\text{upsampled} \in \mathbb{R}^{C \times 2H \times 2W} \). The upsampled feature map is refined through a convolutional block comprising two convolutional layers, each followed by batch normalization (\( \mathcal{B}_N \)) and ReLU activation (\( \sigma \)):
\begin{equation} 
 F_\text{refined} = \sigma \big( \mathcal{B}_N \big( \Gamma_2 \, \sigma \big( \mathcal{B}_N \big( \Gamma_1 F_\text{upsampled} + b_1 \big) \big) + b_2 \big) \big), 
 \label{eq12} 
\end{equation} 
where \( \Gamma_1, \Gamma_2 \) are convolutional kernels and \( b_1, b_2 \) are biases. The number of feature channels is progressively reduced in the decoder stages to enable the reconstruction of spatial details. The final segmentation map is obtained through a \( 1 \times 1 \) convolution, yielding \( F^\prime_0 \in \mathbb{R}^{H_0 \times W_0 \times C_Y} \), where \( C_Y \) represents the number of output channels, corresponding to the ground truth segmentation \( Y \).

\subsection{Loss Functions}
\label{sec3.3}
A customized loss function combining Binary Cross-Entropy (BCE) Loss and Dice Loss is used to ensure accurate and balanced segmentation. BCE provides pixel-wise accuracy, while Dice Loss captures region-level overlap, which is essential to distinguish fine boundaries in tooth segmentation. The BCE loss is defined as:
\begin{equation}
 \text{BCE} = -\frac{1}{N} \sum_{i=1}^N \left[ y_i \log\left(\sigma(x_i)\right) + (1 - y_i)\log\left(1 - \sigma(x_i)\right) \right],
 \label{eq13}
\end{equation}
where $\sigma(x_i)$ is the Sigmoid activation function. The Dice Loss is computed as:
\begin{equation}
 \text{Dice}_{\text{loss}} = 1 - \frac{2 \sum_{i=1}^N (\hat{y}_i \cdot y_i) + \epsilon}{\sum_{i=1}^N \hat{y}_i + \sum_{i=1}^N y_i + \epsilon},
 \label{eq14}
\end{equation}
where $\hat{y}_i = \sigma(x_i)$ is the predicted mask, $y_i$ is the ground truth, and $\epsilon$ is a smoothing constant, set to $1 \times 10^{-5}$ to prevent division by zero. The final loss is given by:
\begin{equation}
 \text{Loss} = \frac{1}{2}\text{BCE} + \text{Dice}_{\text{loss}}
 \label{eq15}
\end{equation}

\section{Experiments}
\label{sec4}

The proposed contextual semantic segmentation network based on KANs with dual encoder is evaluated on dental panoramic x-ray images using two publicly available datasets: the \textit{Children's Dental Panoramic Radiographs Dataset} (CDPR) \cite{51} and the \textit{Teeth Segmentation Dataset} (HTL) \cite{52}. During training, 1500 samples from the CDPR dataset (adult patients) are used, while testing is carried out on 500 samples from the same dataset and a total of 598 samples from the HTL dataset. Augmented images are processed through a ResNet-18 encoder, while nonaugmented images are passed through a customized CNN encoder, both operating on input sizes of $320 \times 320$. All experiments are implemented in PyTorch and executed on a NVIDIA L20 GPU with an Intel Xeon Platinum 8457C processor. 

The model is trained with a batch size of 32 using the Stochastic Gradient Descent (SGD) optimizer with momentum of 0.9 and weight decay of $1 \times 10^{-4}$. A learning rate $1 \times 10^{-4}$ is selected after preliminary experiments with a search space of {$1 \times 10^{-3}, 1 \times 10^{-4}, 1 \times 10^{-5}$}, as it provided the best balance between convergence speed and stable optimization. A Cosine Annealing learning rate scheduler with a minimum learning rate of $1\times10^{-5}$ is used to further stabilize the optimization process. During training, 80\% of the training data are used to train the model and the rest 20\% is used for validation. The network is trained for 200 epochs with early stopping based on validation loss to prevent overfitting.

\subsection{Evaluation Metrics}
\label{sec4.1}
To evaluate the performance of the proposed model for the segmentation of dental images, we employ four commonly used metrics: mean Intersection over Union (\textit{mIoU}), \textit{Dice Coefficient}, \textit{Accuracy}, and \textit{Recall}. These metrics provide a comprehensive assessment of segmentation quality by comparing the overlap between predicted and ground truth masks. The mathematical expressions for the evaluation metrics are as follows:
\begin{equation}
  \textit{mIoU} = \frac{1}{C} \sum_{i=1}^C \frac{\text{TP}_i}{\text{TP}_i + \text{FP}_i + \text{FN}_i}
  \label{eq16}
\end{equation}
\begin{equation}
  \textit{Dice}_{coefficient} = \frac{2 \sum_{i=1}^N (\hat{y}_i \cdot y_i) + \epsilon}{\sum_{i=1}^N \hat{y}_i + \sum_{i=1}^N y_i + \epsilon}
  \label{eq17}
\end{equation}
\begin{equation}
  \textit{Accuracy} = \frac{TP + TN}{TP + TN + FP + FN}
  \label{eq18}
\end{equation}
\begin{equation}
  \textit{Recall} = \frac{TP}{TP + FN}
  \label{eq19}
\end{equation}
Here, TP, FP, FN and TN represent True Positives, False Positives, False Negatives, and True Negatives, respectively. In Equation~\ref{eq16}, TP, FP and FN are calculated for each class $i$, and $C$ is the number of total classes. Equation~\ref{eq17} defines \textit{Dice Coefficient}, calculating the harmonic mean of precision and recall, where $\hat{y}_i$ and $y_i$ are the predicted values of the pixel and ground truth, respectively, and $\epsilon$ prevents division by zero. In Equation~\ref{eq18}, \textit{Accuracy} measures the proportion of correctly predicted pixels of the total pixels. Equation~\ref{eq19} defines \textit{Recall}, evaluating the model's ability to correctly identify positive pixels, calculated as the ratio of $TP$ to $TP + FN$.

\subsection{Results}
\label{sec4.2}
\subsubsection{Empirical analysis}
\label{sec4.2.1}

\begin{figure}[h]
  \centering
  \begin{minipage}[c]{1\textwidth}
    \begin{minipage}[c]{0.03\textwidth}
      \raggedleft
      {\scriptsize\text{(a)}}
    \end{minipage}
    \begin{subfigure}[c]{0.15\textwidth}
      \centering
      \includegraphics[width=\textwidth]{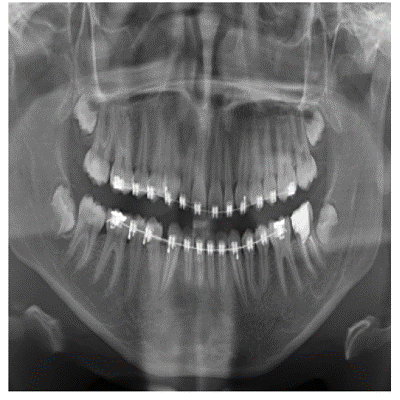}
    \end{subfigure}
    \begin{subfigure}[c]{0.15\textwidth}
      \centering
      \includegraphics[width=\textwidth]{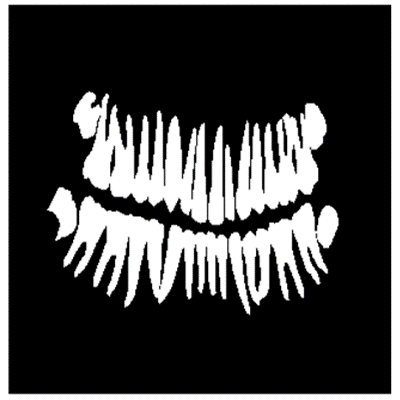}
    \end{subfigure}
    \begin{subfigure}[c]{0.15\textwidth}
      \centering
      \includegraphics[width=\textwidth]{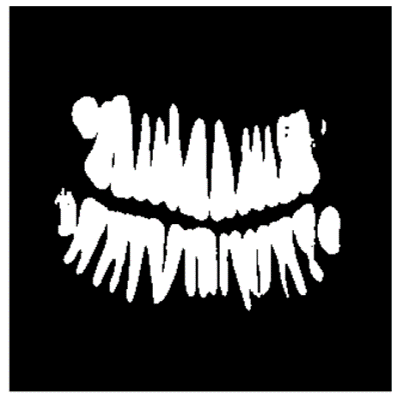}
    \end{subfigure}
    \begin{subfigure}[c]{0.15\textwidth}
      \centering
      \includegraphics[width=\textwidth]{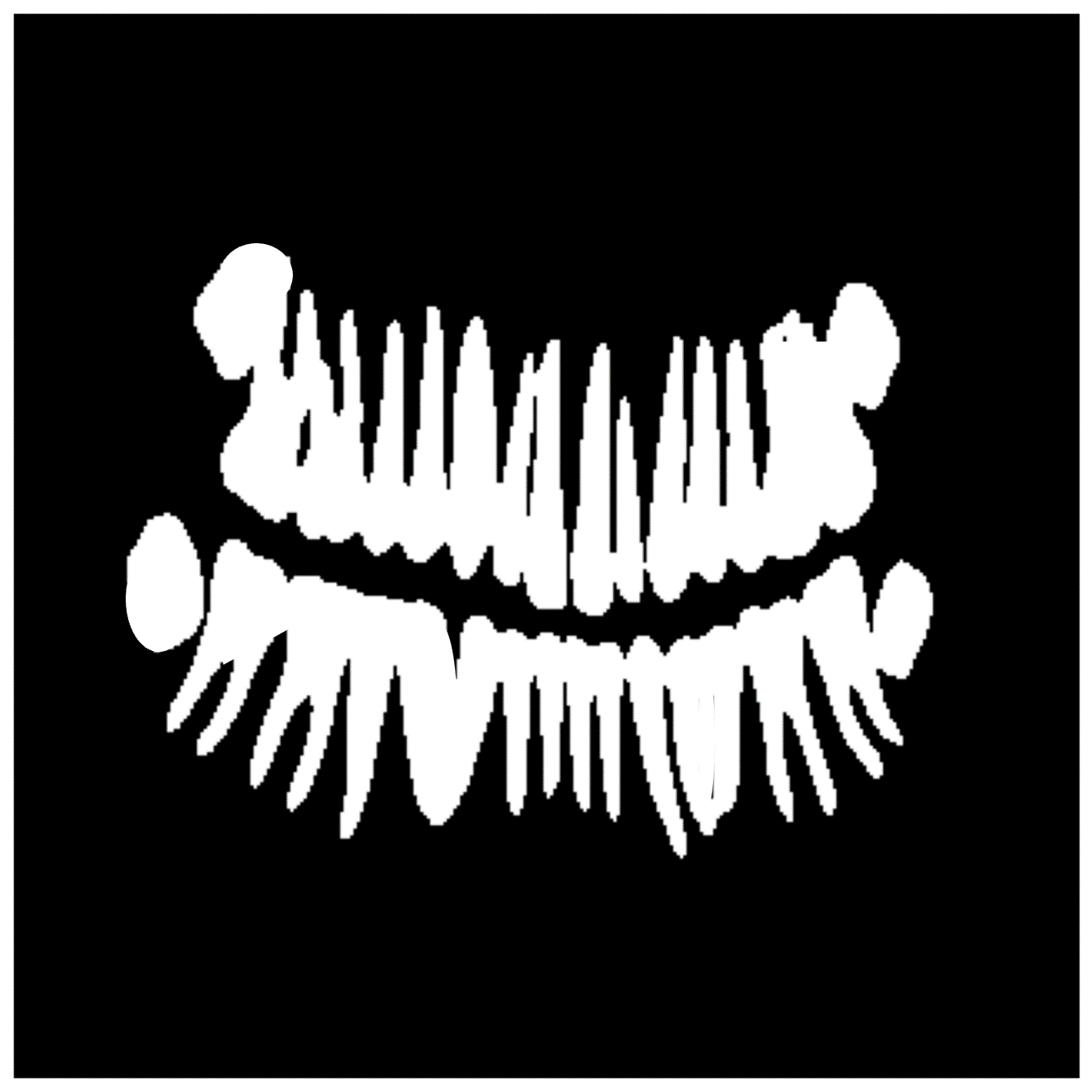}
    \end{subfigure}
    \begin{subfigure}[c]{0.15\textwidth}
      \centering
      \includegraphics[width=\textwidth]{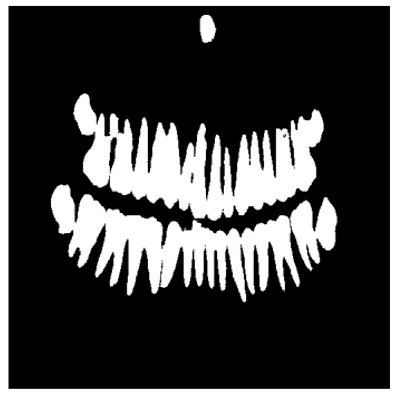}
    \end{subfigure}
    \begin{subfigure}[c]{0.15\textwidth}
      \centering
      \includegraphics[width=\textwidth]{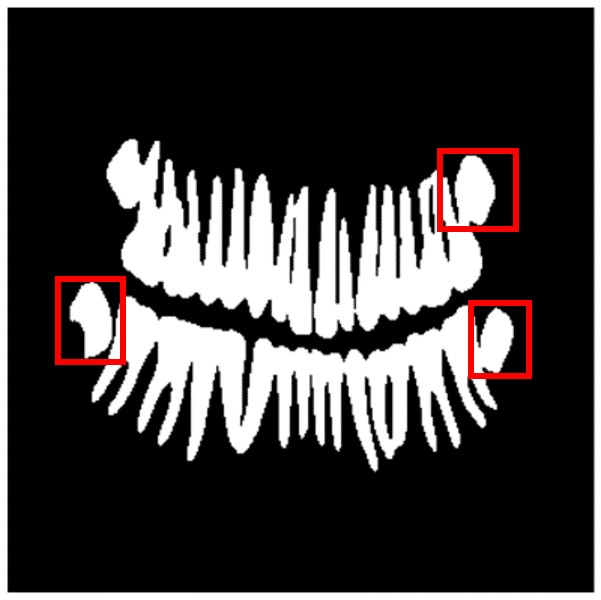}
    \end{subfigure}

    \vspace{0.5em}
    \begin{minipage}[c]{0.03\textwidth}
      \raggedleft
      {\scriptsize\text{(b)}}
    \end{minipage}
    \begin{subfigure}[c]{0.15\textwidth}
      \centering
      \includegraphics[width=\textwidth]{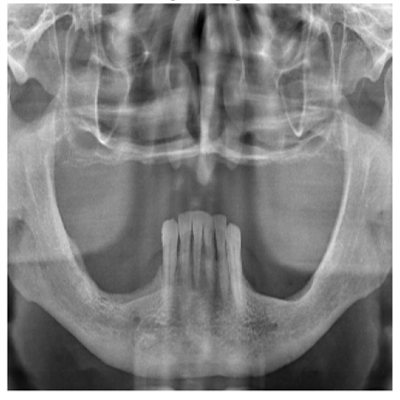}
    \end{subfigure}
    \begin{subfigure}[c]{0.15\textwidth}
      \centering
      \includegraphics[width=\textwidth]{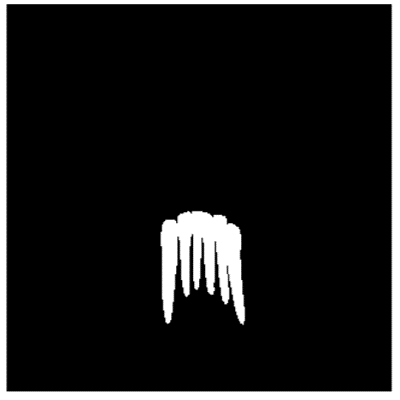}
    \end{subfigure}
    \begin{subfigure}[c]{0.15\textwidth}
      \centering
      \includegraphics[width=\textwidth]{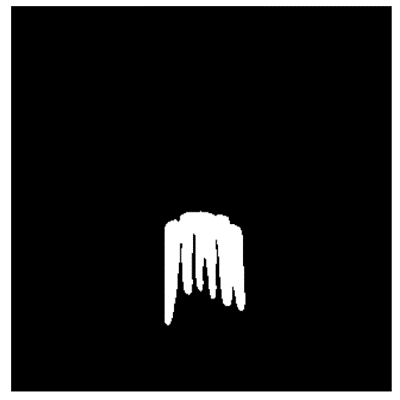}
    \end{subfigure}
    \begin{subfigure}[c]{0.15\textwidth}
      \centering
      \includegraphics[width=\textwidth]{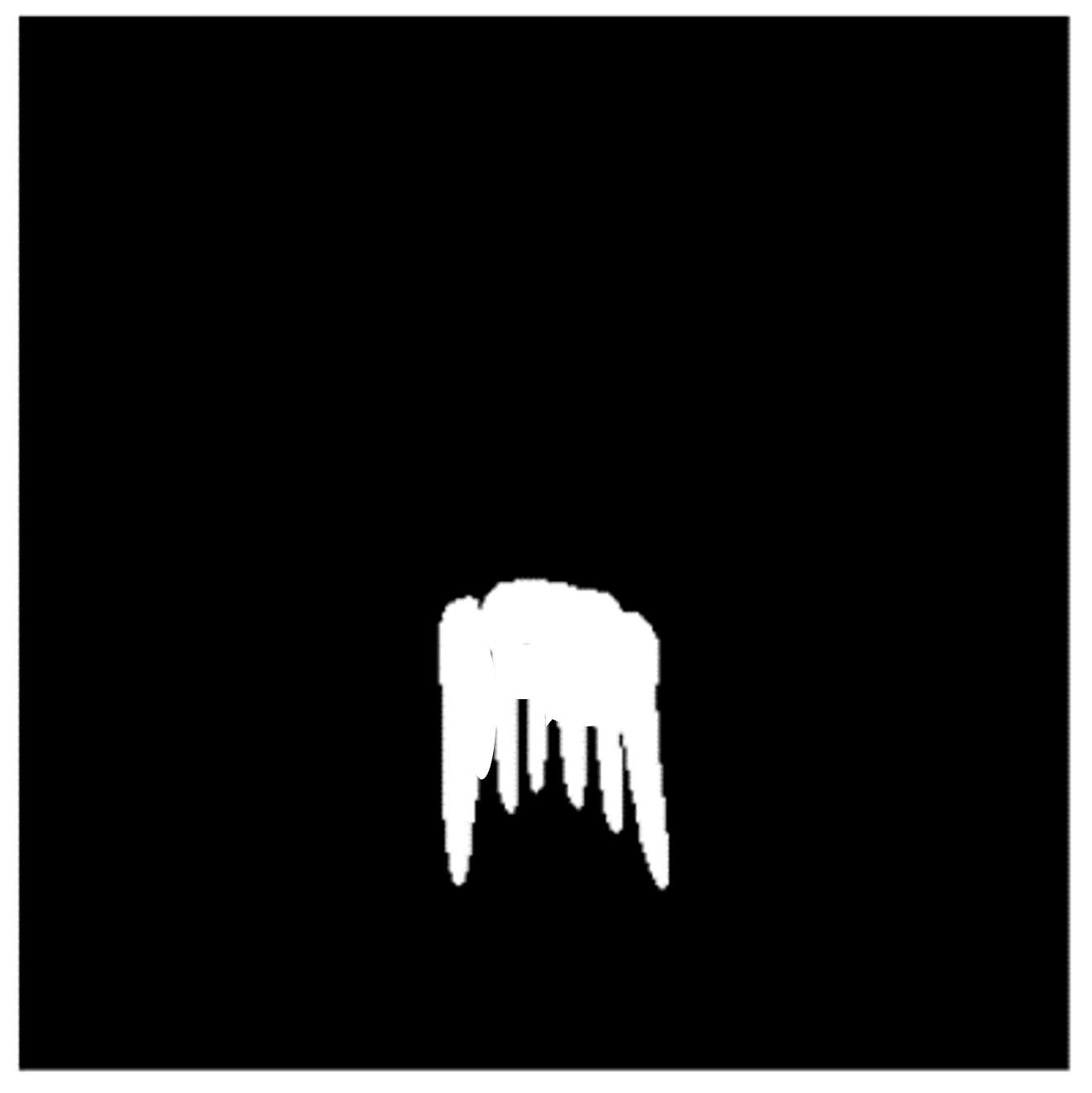}
    \end{subfigure}
    \begin{subfigure}[c]{0.15\textwidth}
      \centering
      \includegraphics[width=\textwidth]{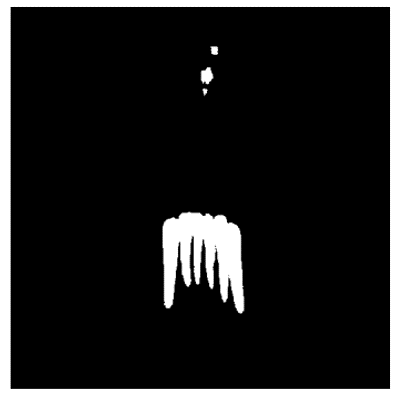}
    \end{subfigure}
    \begin{subfigure}[c]{0.15\textwidth}
      \centering
      \includegraphics[width=\textwidth]{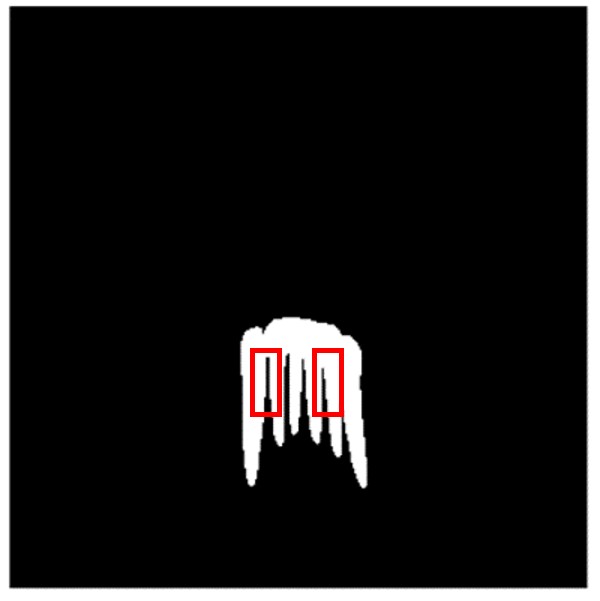}
    \end{subfigure}

    \vspace{0.5em}
    \begin{minipage}[c]{0.03\textwidth}
      \raggedleft
      {\scriptsize\text{(c)}}
    \end{minipage}
    \begin{subfigure}[c]{0.15\textwidth}
      \centering
      \includegraphics[width=\textwidth]{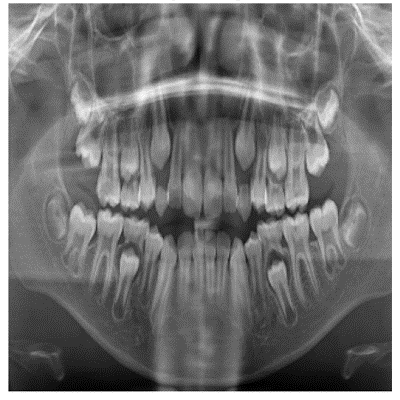}
    \end{subfigure}
    \begin{subfigure}[c]{0.15\textwidth}
      \centering
      \includegraphics[width=\textwidth]{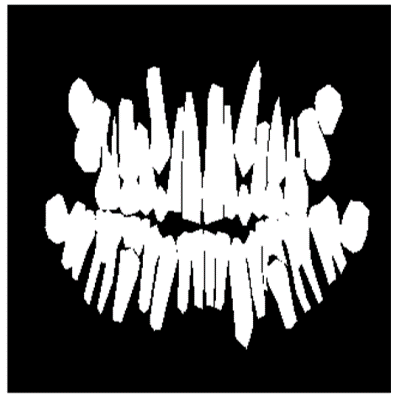}
    \end{subfigure}
    \begin{subfigure}[c]{0.15\textwidth}
      \centering
      \includegraphics[width=\textwidth]{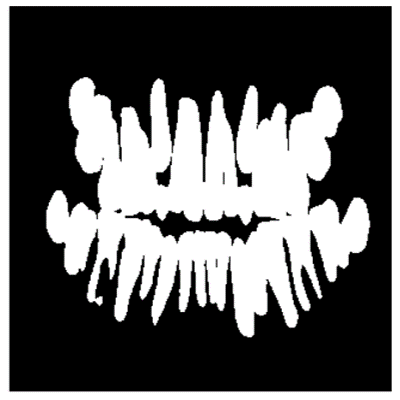}
    \end{subfigure}
    \begin{subfigure}[c]{0.15\textwidth}
      \centering
      \includegraphics[width=\textwidth]{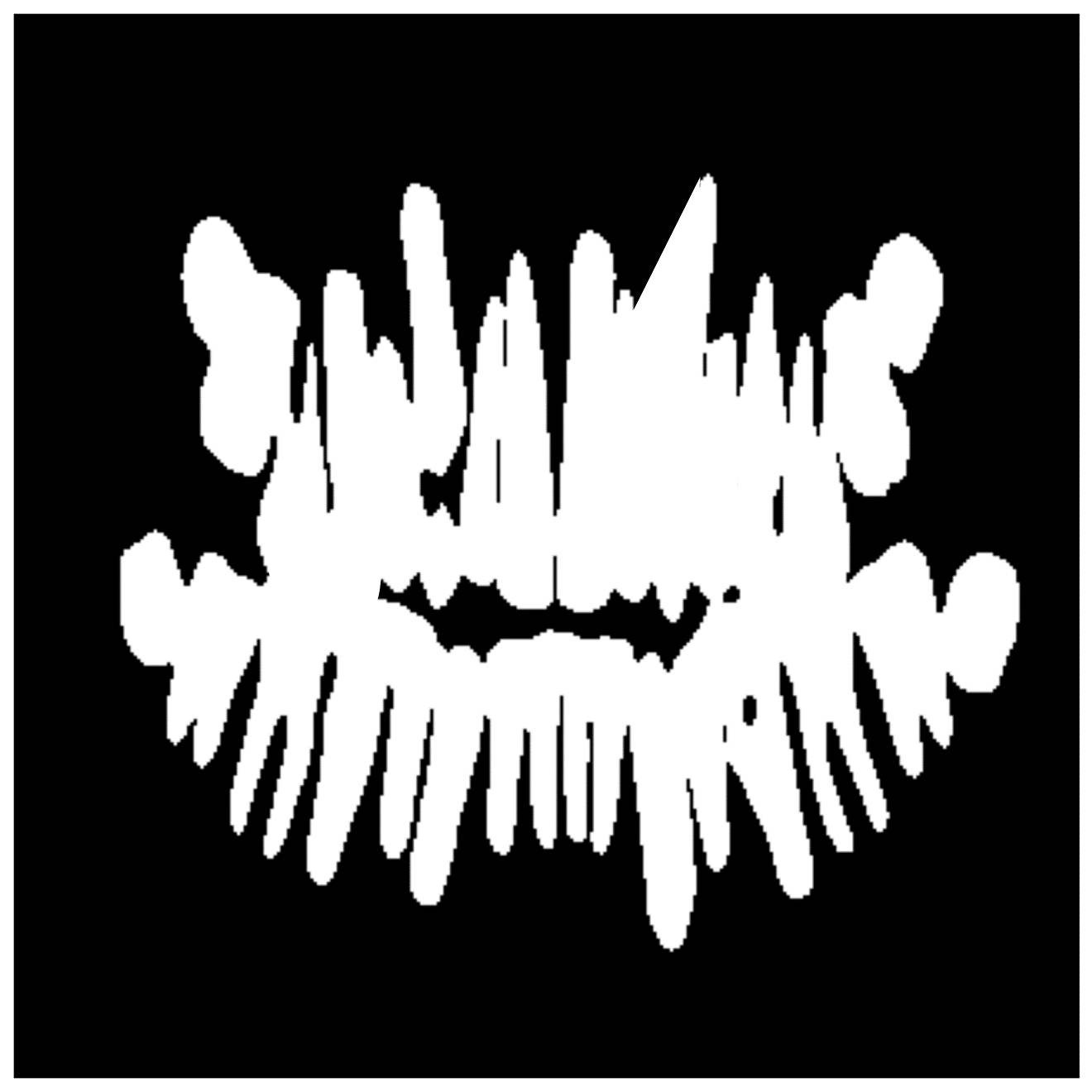}
    \end{subfigure}
    \begin{subfigure}[c]{0.15\textwidth}
      \centering
      \includegraphics[width=\textwidth]{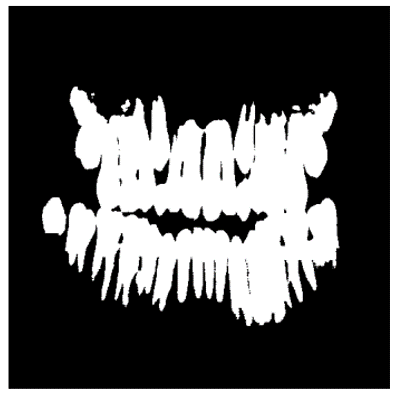}
    \end{subfigure}
    \begin{subfigure}[c]{0.15\textwidth}
      \centering
      \includegraphics[width=\textwidth]{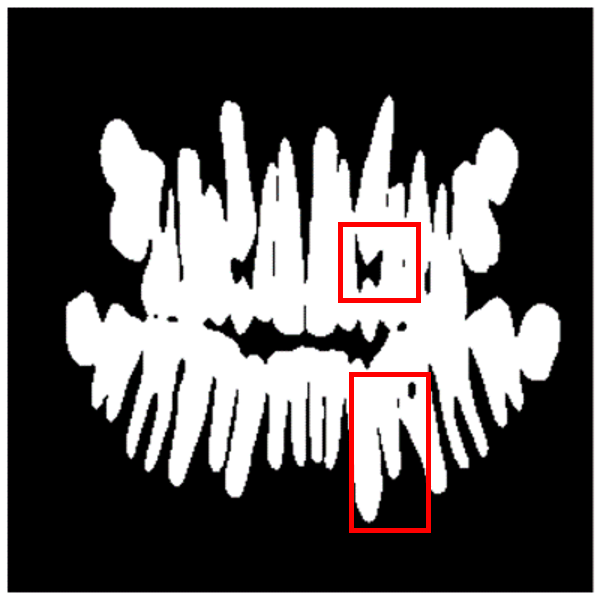}
    \end{subfigure}

    \vspace{0.5em}
    \begin{minipage}[c]{0.03\textwidth}
      \raggedleft
      {\scriptsize\text{(d)}}
    \end{minipage}
    \begin{subfigure}[c]{0.15\textwidth}
      \centering
      \includegraphics[width=\textwidth]{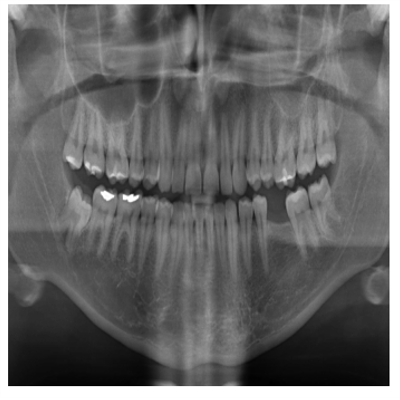}
      \caption*{Original image}
    \end{subfigure}
    \begin{subfigure}[c]{0.15\textwidth}
      \centering
      \includegraphics[width=\textwidth]{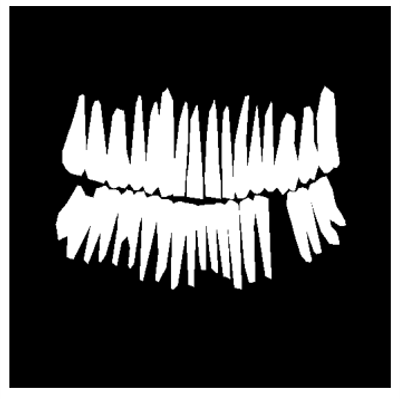}
      \caption*{Ground Truth}
    \end{subfigure}
    \begin{subfigure}[c]{0.15\textwidth}
      \centering
      \includegraphics[width=\textwidth]{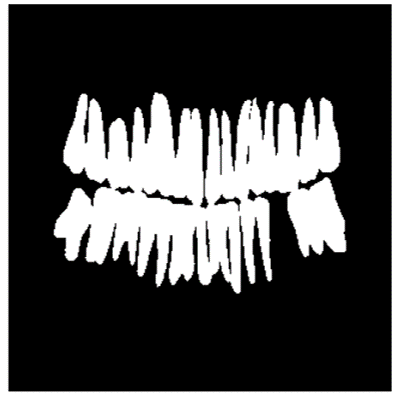}
      \caption*{U-KAN\cite{46}}
    \end{subfigure}
    \begin{subfigure}[c]{0.15\textwidth}
      \centering
      \includegraphics[width=\textwidth]{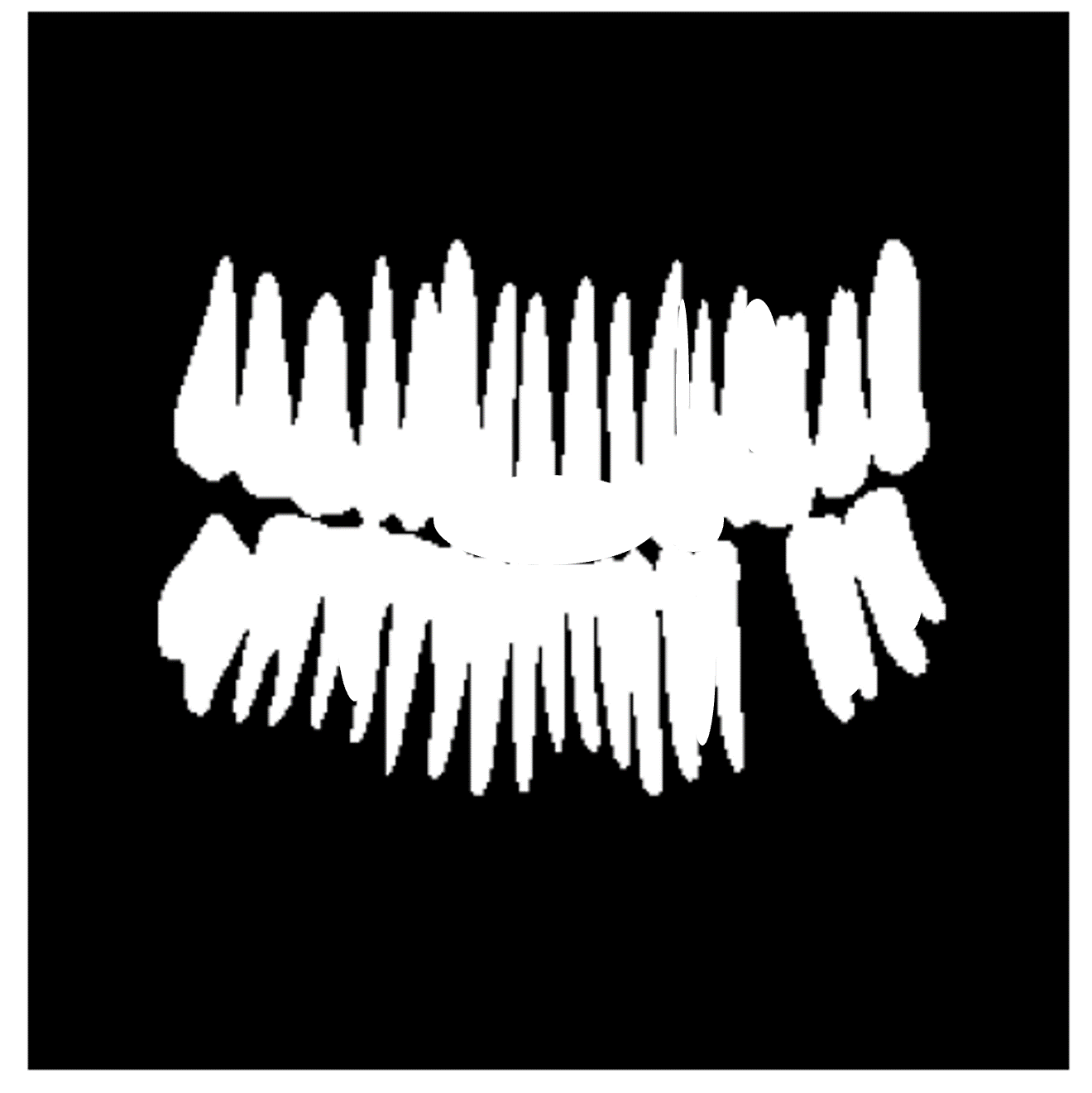}
      \caption*{YOLOv8\cite{53}}
    \end{subfigure}
    \begin{subfigure}[c]{0.15\textwidth}
      \centering
      \includegraphics[width=\textwidth]{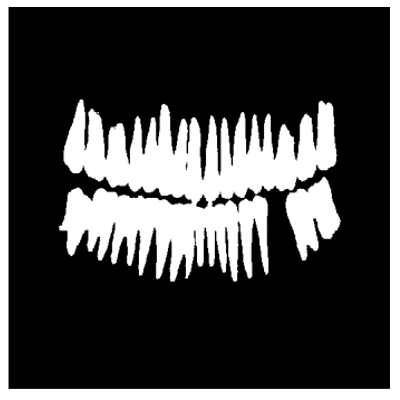}
      \caption*{TeethU-Net\cite{27}}
    \end{subfigure}
    \begin{subfigure}[c]{0.15\textwidth}
      \centering
      \includegraphics[width=\textwidth]{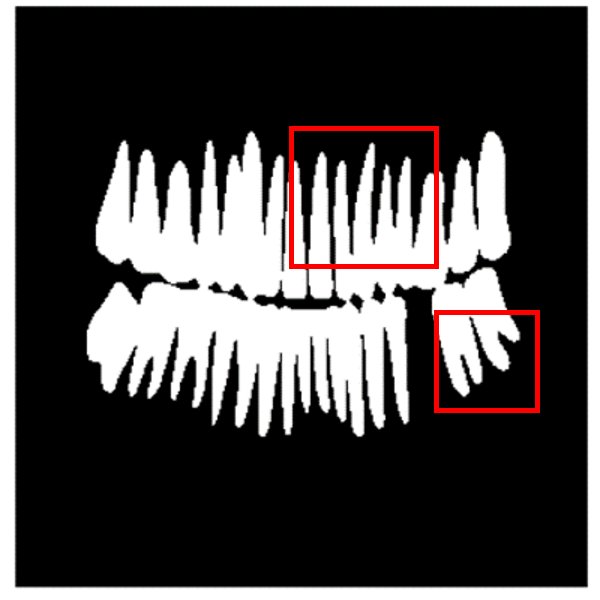}
      \caption*{Our method}
    \end{subfigure}
  \end{minipage}
  \caption{Qualitative Comparison of our proposed network with existing state-of-the-art networks.}
  \label{fig:4}
\end{figure}

We conduct extensive experiments on state-of-the-art models, including U-KAN \cite{46}, YOLOv8 \cite{53}, and Teeth U-Net \cite{27}, to validate the effectiveness of our proposed DE-KAN model. U-KAN \cite{44} excels in medical 2D image segmentation; YOLOv8 \cite{53}, designed primarily for object detection, also performs well in segmentation tasks; and Teeth U-Net \cite{27} is specifically designed for dental image segmentation. For a fair comparison, all models are trained and evaluated under identical settings on two benchmark dental panoramic radiograph datasets: CDPR \cite{51} and HTL \cite{52}. We augment 50\% of the images during training, leaving the rest non-augmented from the CDPR dataset \cite{51}, since all these models contain a single encoder. We do not apply augmentation during testing. Qualitative and quantitative results are presented in Figure~\ref{fig:4} and Table~\ref{table:1},\ref{table:2} respectively. Tables~\ref{table:1} and \ref{table:2} demonstrate that DE-KAN consistently outperforms other models across both CDPR\cite{51} and HTL\cite{52} dataset respectively. On the CDPR dataset\cite{51}, DE-KAN achieves the highest mIoU of 94.5\% and a Dice Coefficient of 97.10\%. Similarly, on the HTL dataset\cite{52}, it achieves an mIoU of 89.45\% and a Dice Coefficient of 94.39\%.

\begin{table}[h] 
  \begin{center}
  \caption{Comparison with state-of-the-art segmentation models on CDPR datasets\cite{51}. Bold text indicates the best result for each metric.
  \label{table:1}}
  \resizebox{0.9\textwidth}{!}{
  \begin{tabular}{lcccc}
   \toprule
    Methods & mIoU(\%) & Dice Coefficient(\%) & Accuracy(\%) & Recall(\%) \\ 
    \midrule
      U-KAN \cite{46} & 86.50 & 91.79 & 95.95 & 92.20 \\
      Teeth U-Net \cite{27} & 87.96 & 92.21 & 96.09 & 92.78 \\
      YOLOv8 \cite{53} & 85.91 & 90.90 & 93.30 & 91.68 \\
      Our (DE-KAN) & \textbf{94.5} & \textbf{97.10} & \textbf{98.91} & \textbf{97.36} \\
      \bottomrule
    \end{tabular}
    }
    \end{center}
\end{table}

\begin{table}[h] 
  \begin{center}
  \caption{Comparison with state-of-the-art segmentation models on HTL datasets\cite{52}. Bold text indicates the best result for each metric.
  \label{table:2}}
  \resizebox{0.9\textwidth}{!}{
  \begin{tabular}{lcccc}
   \toprule
    Methods & mIoU(\%) & Dice Coefficient(\%) & Accuracy(\%) & Recall(\%) \\ 
    \midrule
      U-KAN \cite{46} & 83.79 & 89.30 & 89.67 & 87.30 \\
      Teeth U-Net \cite{27} & 84.27 & 89.69 & 90.07 & 88.50 \\
      YOLOv8 \cite{53} & 82.14 & 89.47 & 88.90 & 87.65 \\
      Our (DE-KAN) & \textbf{89.45} & \textbf{94.39} & \textbf{94.91} & \textbf{93.45} \\
      \bottomrule
    \end{tabular}
    }
    \end{center}
\end{table}

\begin{table}[h] 
  \begin{center}
  \caption{Computational comparison with state-of-the-art segmentation models. The arrow $\downarrow$ means a smaller value is better, while $\uparrow$ shows a larger value is better.
  \label{table:3}}
  \resizebox{0.9\textwidth}{!}{
  \begin{tabular}{lcccc}
   \toprule
    Methods & GFLOPs $\downarrow$ & Parameters(M) $\downarrow$ & Latency (ms) $\downarrow$ & Throughput (FPS) $\uparrow$\\ 
    \midrule
      U-KAN \cite{46} & 16.44 & 25.36 & 37.82 & 26.44 \\
      Teeth U-Net \cite{27} & 75.24 & 36.04 & 55.63 & 17.98 \\
      YOLOv8 \cite{53} & 26.20 & 36.90 & 44.27 & 22.59 \\
      Our (DE-KAN) & 164.42 & 32.73 & 59.05 & 16.93 \\
      \bottomrule
    \end{tabular}
    }
    \end{center}
\end{table}


Compared to U-KAN\cite{46}, which integrates four tokenized KAN blocks into a U-Net backbone, DE-KAN adopts a dual encoder method paired with two KAN blocks. This design leverages local and global features more effectively, enhancing feature extraction and segmentation accuracy. This DE-KAN configuration outperforms U-KAN, particularly in handling complex spatial hierarchies and detailed pixel-level predictions. Our model also surpasses YOLOv8\cite{53}, which employs a mix of traditional and advanced CNNs, achieving significant improvements in \textit{mIoU} and \textit{Dice Coefficient} across both datasets. Furthermore, DE-KAN outperforms Teeth U-Net\cite{27}, providing substantial accuracy gains. For example, on the CDPR dataset\cite{51}, DE-KAN records a $6.54\%$ increase in \textit{mIoU} and a $4.89\%$ rise in \textit{Dice Coefficient}. On the HTL data set\cite{52}, it achieves an improvement of $5.18\%$ and $5.09\%$ improvement in \textit{mIoU} and \textit{Dice Coefficient}, respectively. As shown in Figure~\ref{fig:4}, a visual comparison demonstrates the superior segmentation performance at the pixel level of our method on the CDPR dataset \cite{51}. Our approach reduces pixel-level misdetection of overlapping teeth and improves segmentation accuracy, particularly along the sharp edges of the teeth and the jawbone.

Although DE-KAN delivers superior segmentation performance, it comes with only one trade-off in computational cost due to its complex architecture, as stated in Table~\ref{table:3}. However, this higher computational requirement enables the model to better address the intricate structure of dental objects, achieving significantly improved accuracy compared to models that are less computationally intensive. Notably, despite the increased complexity, the latency remains acceptable (59.05 ms per image), only marginally higher than lighter models, ensuring practical applicability without sacrificing efficiency.

\subsubsection{Comparison with other state-of-the-art networks on CDPR dataset\cite{51}}
\label{sec4.2.2}
\begin{table}[h] 
  \centering
  \caption{Comparison with other state-of-the-art methods on CDPR dataset \cite{51}. Bold text indicates the best result for each metric.
  \label{table:4}}
  \resizebox{0.9\textwidth}{!}{
  \begin{tabular}{lcccc}
   \toprule
    Methods & mIoU(\%) & Dice Coefficient(\%) & Accuracy(\%) & Recall(\%) \\ 
    \midrule
  U-Net \cite{unet} & 88.58\% & 93.92\% & 97.19\% & 94.59\% \\
  R2 U-Net \cite{54} & 88.92\% & 94.11\% & 97.24\% & 93.51\% \\
  PSPNet \cite{55} & 86.93\% & 92.99\% & 96.70\% & 91.64\% \\
  Deeplab V3+ \cite{56} & 86.39\% & 92.67\% & 96.65\% & 94.65\% \\
  Our (DE-KAN)  & \textbf{94.5\%} & \textbf{97.1\%} & \textbf{98.91\%} & \textbf{97.36\%} \\
  \bottomrule
  \end{tabular}
  }
 \end{table}

 As reported by Zhang et al.\cite{51}, the results in Table~\ref{table:4} demonstrate that our proposed DE-KAN model achieves superior performance compared to other state-of-the-art methods, obtaining the highest scores across all metrics. Specifically, DE-KAN achieves an \textit{mIoU} of 94.5\%, \textit{Accuracy} of 98.91\%, \textit{Recall} of 97.36\%, and a Dice Coefficient of 97.1\%, surpassing traditional models such as U-Net \cite{unet}, R2 U-Net \cite{54}, PSPNet \cite{55}, and DeepLab V3+ \cite{56}. These findings validate the effectiveness of DE-KAN for precise segmentation on the CDPR dataset\cite{51}.

\subsection{Ablation study}
\label{sec4.3}
\begin{table}[htbp] 
  \centering
  \caption{Comparison of ablation methods on CDPR dataset. Bold text indicates the best result for each metric. The arrow $\downarrow$ means a smaller value is better, while $\uparrow$ shows a larger value is better.
  \label{table:5}}
  \resizebox{\textwidth}{!}
   {\renewcommand{\arraystretch}{1.3} 
  \begin{tabular}{lcccccc}
   \toprule
    {Methods} & {mIoU}(\%)$\uparrow$ & {Dice Coefficient}(\%)$\uparrow$ & {Accuracy}(\%)$\uparrow$ & {Recall}(\%)$\uparrow$ & {Parameters (M)}$\downarrow$ & {GFLOPs}$\downarrow$ \\ 
    \midrule
  Baseline (No KANs) & 53.34\% & 68.56\% & 86.72\% & 78.84\% & 23.21 & 164.01 \\
  KANs + CNN encoder & 64.74\% & 78.31\% & 91.45\% & 82.71\% & \textbf{21.56} & 160.87 \\
  KANs + ResNet-18 encoder    & 55.39\% & 70.50\% & 88.67\% & 75.50\% & 31.18 & \textbf{5.76} \\
  Our (DE-KAN)  & \textbf{94.5\%} & \textbf{97.1\%} & \textbf{98.91\%} & \textbf{97.36\%} & 32.73 & 164.42 \\
    \bottomrule
   \end{tabular}}
\end{table}

\begin{figure}[h]
  \centering
  \includegraphics[width=0.95\linewidth]{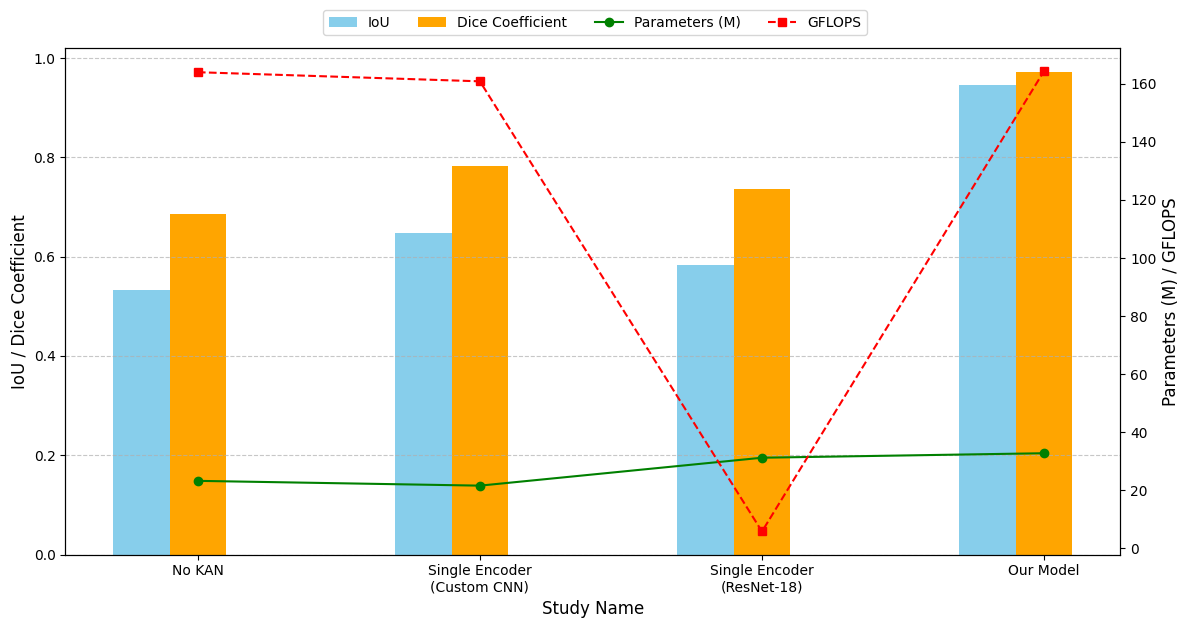}
  \caption{Visualization of ablation study comparing segmentation accuracy and computational efficiency across different network configurations based on Table~\ref{table:5}.
  \label{fig:5}}
\end{figure}

We perform an ablation study to assess the contribution of individual components in our proposed Dual Encoder KAN (DE-KAN) architecture. The quantitative results are presented in Table~\ref{table:5} and visualized in Figure~\ref{fig:5}. The baseline model, without KAN blocks, demonstrates the lowest segmentation performance, achieving an \textit{mIoU} of 53.34\% and a \textit{Dice Coefficient} of 68.56\%. Although it has fewer parameters (23.21M), it exhibits a high computational cost (164.01 GFLOPs), highlighting inefficiencies in both accuracy and computational performance. To isolate the impact of each encoder, we evaluate two single-encoder variants:
\begin{enumerate}
\item {KANs with CNN Encoder:} Incorporating KAN blocks with only the CNN encoder improves performance to an \textit{mIoU} of 64.74\% and a \textit{Dice Coefficient} of 78.31\%. This configuration reduces both the number of parameters (21.56M) and the computational cost (160.87 GFLOPs) compared to the baseline model.
\item {KANs with ResNet-18 Encoder:} Replacing the CNN encoder with a ResNet-18 backbone yields marginal improvements (\textit{mIoU}: 55.39\%, \textit{Dice Coefficient}: 70.50\%), compared to the baseline model. However, this variant significantly reduces computational cost (5.76 GFLOPs) while increasing the number of parameters to 31.18M, reflecting a trade-off between performance and efficiency.
\end{enumerate}

Our proposed \textbf{DE-KAN} architecture, combining a CNN encoder and a ResNet-18 backbone with KAN blocks, achieves state-of-the-art performance on all metrics (\textit{mIoU}: 94.5\%, \textit{Dice Coefficient}: 97.1\%, \textit{Accuracy}: 98.91\%, and \textit{Recall}: 97.36\%). Despite a modest increase in the number of parameters (32.73M), the GFLOPs (164.42) remain comparable to the baseline. These results highlight the complementary benefits of dual encoder design and the effectiveness of KAN modules in improving segmentation performance. In conclusion, while single encoder variants offer incremental improvements over the baseline, they remain limited in segmentation accuracy and computational efficiency. Our DE-KAN architecture effectively balances accuracy, recall, and computational cost, delivering superior performance on the CDPR dataset\cite{51}.

\subsection{Discussion}
\label{sec4.4}
Teeth are vital organs directly connected to the human brain and accurate segmentation is critical for the early diagnosis and treatment of dental conditions. Compared to existing state-of-the-art models, our proposed model achieves superior segmentation accuracy, aiding radiologists and dentists in early, precise patient diagnoses, and potentially saving lives. Furthermore, our model extends beyond tooth segmentation and is promising for dentistry-related tasks such as detecting dental cavities and identifying missing teeth.

The complex structure of our model results in higher computational demands compared to existing models. Although this complexity improves the ability of the network to segment intricate dental structures, it requires substantial computational resources for real-world application. However, this trade-off enables our model to achieve exceptional segmentation accuracy, providing dentists with deeper insights for diagnosis and treatment planning.
\subsection{Future Work}
\label{sec4.5}
This study offers an optimistic direction for future research. \textit{First}, this study focuses solely on teeth segmentation from panoramic radiographic 2D images. This study has significant potential to handle images from various modalities. In pursuit of this goal, our future work will emphasize minimizing the limitations of the modality. \textit{Second}, this study only addresses 2-dimensional images for teeth segmentation; our future work will be extended to 3-dimensional images.

\section{Conclusion}
\label{sec5}
This study introduces the Dual Encoder KAN (DE-KAN) model, designed to improve feature extraction and segmentation in complex image datasets. During training, the model integrates a ResNet-18 backbone and a customized CNN encoder to capture hierarchical abstract features and local spatial information from augmented and nonaugmented input. We exclusively apply augmentation during training to improve generalization, prevent overfitting, and increase data diversity. During testing, making the inputs into both encoders the same ensures reliable performance in practical applications. This enables the model to handle a wide range of conditions while keeping its accuracy and robustness during deployment. Empirical evaluations on two benchmark datasets demonstrate the effectiveness of DE-KAN in addressing the challenges of dentition segmentation in medical images. In general, we believe that the use of dual encoders with KANs has the potential of unconventional designs to enhance performance in diverse visual tasks, particularly in medical imaging applications.

\bibliographystyle{elsarticle-num} 
\bibliography{./DE-KAN}



\end{document}